\documentclass[10pt,twocolumn,letterpaper]{article}

\usepackage{cvpr}              % To produce the CAMERA-READY version
\usepackage{stmaryrd}
\usepackage{fix-cm}
\usepackage{multirow}
\usepackage{threeparttable}
\usepackage{colortbl}
\usepackage{booktabs}
\usepackage{caption}
\usepackage{subcaption}
\usepackage{colortbl}
\usepackage[accsupp]{axessibility}

\usepackage{stmaryrd}
\newcommand{\red}[1]{{\color{red}#1}}

\definecolor{cvprblue}{rgb}{0.21,0.49,0.74}
\usepackage[pagebackref,breaklinks,colorlinks,allcolors=cvprblue]{hyperref}

\def\paperID{7883} % *** Enter the Paper ID here
\def\confName{CVPR}
\def\confYear{2025}

\title{Probabilistic Prompt Distribution Learning for Animal Pose Estimation}

\author{Jiyong Rao$^1$\quad Brian Nlong Zhao$^{2\dag}$\quad Yu Wang$^1$\thanks{Corresponding author.} \\
$^1$~School of Computer Science and Technology, Tongji University\quad $^2$~Stanford University\\
{\tt\small jiyongrao@tongji.edu.cn},\quad {\tt\small briannlz@stanford.edu},\quad {\tt\small yuwangtj@yeah.net}
}

\begin{document}
\maketitle
\def\thefootnote{\dag}\footnotetext{Equal contribution.}
\begin{abstract}
Multi-species animal pose estimation has emerged as a challenging yet critical task, hindered by substantial visual diversity and uncertainty. This paper challenges the problem by efficient prompt learning for Vision-Language Pretrained (VLP) models, \textit{e.g.} CLIP, aiming to resolve the cross-species generalization problem. At the core of the solution lies in the prompt designing, probabilistic prompt modeling and cross-modal adaptation, thereby enabling prompts to compensate for cross-modal information and effectively overcome large data variances under unbalanced data distribution. To this end, we propose a novel probabilistic prompting approach to fully explore textual descriptions, which could alleviate the diversity issues caused by long-tail property and increase the adaptability of prompts on unseen category instance. Specifically, we first introduce a set of learnable prompts and propose a diversity loss to maintain distinctiveness among prompts, thus representing diverse image attributes. Diverse textual probabilistic representations are sampled and used as the guidance for the pose estimation. Subsequently, we explore three different cross-modal fusion strategies at spatial level to alleviate the adverse impacts of visual uncertainty. Extensive experiments on multi-species animal pose benchmarks show that our method achieves the state-of-the-art performance under both supervised and zero-shot settings.
The code is available at \url{https://github.com/Raojiyong/PPAP}.
\end{abstract}    
\section{Introduction}
\label{sec:intro}
Animal Pose Estimation (APE)~\citep{cao2019cross,mathis2018deeplabcut,pereira2019fast,pereira2022sleap,ye2024superanimal} represents a fundamental vision task, the objective of which is to identify a series of keypoint locations in an animal skeleton.
The skeleton (\ie a set of connected coordinates), serves as a minimal yet informative unit for encoding both static structure and motion cues, thereby providing critical support for understanding animal behaviour.
Consequently, APE plays a significant role in various research fields, such as neuroscience~\citep{mathis2018deeplabcut,pereira2019fast}, ecology~\citep{davies2015keep,jiang2022animal}, zoology study~\citep{lauer2022multi,pereira2022sleap} and biotechnological quests~\citep{ng2022animal,weinreb2024keypoint}.

In numerous instances \citep{yu2021apk, ng2022animal, mathis2018deeplabcut}, researchers have opted to directly apply Human Pose Estimation (HPE) methodologies to animals. However, it is evident that many animal species exhibit pose distributions that deviate significantly from human-specific models, leading to substantial domain shifts. To address the multi-species Animal Pose Estimation (APE) challenge, Category-Agnostic Pose Estimation (CAPE) methods \citep{xu2022pose,ren2024dynamic} have been investigated. These methods frame keypoint prediction as a keypoint matching problem, necessitating supplementary support sets that belong to the same category as the test image. While promising, this approach introduces additional computational overhead and necessitates prior knowledge of the animal category, thereby limiting its practicality. Moreover, given the substantial data variances observed across species \citep{yu2021apk,ng2022animal}, vision-only APE methods falter when confronted with cross-species challenges that rely solely on visual cues.

\begin{figure}[t]
  \centering
   \includegraphics[width=1.0\linewidth]{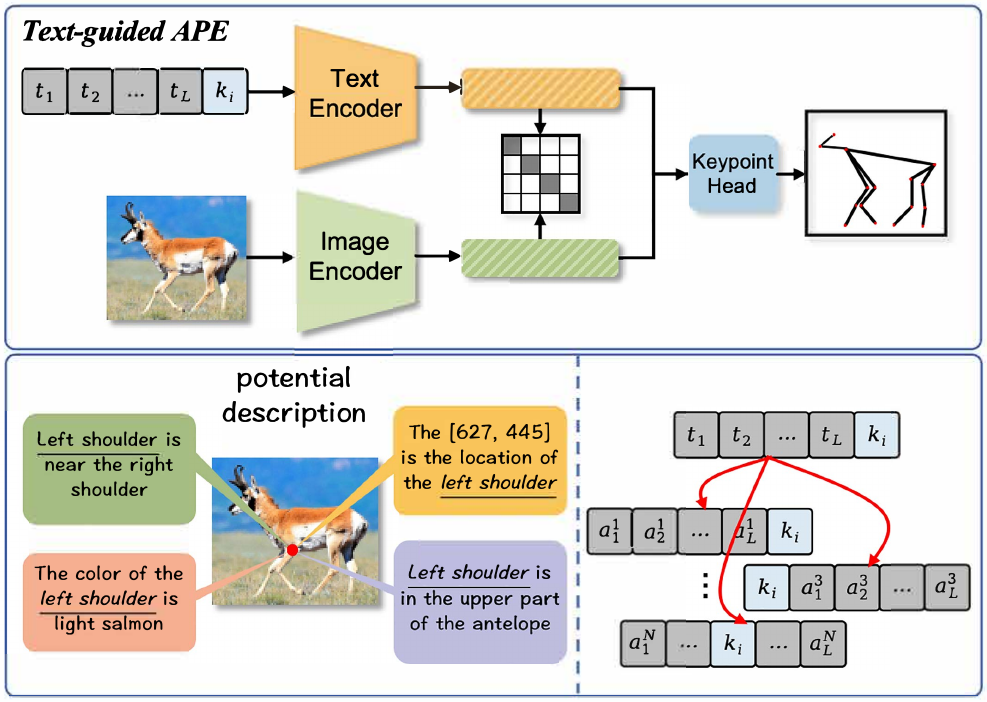}
   \vspace{-10pt}  
   \caption{The motivation that one key point can be complementarily described in different views (An example of left shoulder).}
   \vspace{-15pt}
   \label{fig:intro}
\end{figure}

Recently, the multimodal system proposed in \citep{radford2021clip} has garnered substantial research attention due to its ability to combine complementary data to enhance visual recognition. Despite the distinct visual characteristics of various animal species, they share common textual descriptions, as noted in \citep{zhang2023clamp}. Consequently, the straightforward approach of setting the keypoint name as the class token, known as deterministic prompt learning, has become prevalent in the current literature \citep{zhang2023clamp,yang2025xpose}. For instance, X-Pose \citep{yang2025xpose} utilizes a fixed text template like ``\textit{A [Image Style] photo of a [Keypoint]}", while CLAMP \citep{zhang2023clamp} employs a continuous prompt template of ``\textit{[Learnable Tokens] + [Keypoint]}". However, existing multimodal APE methods fall short in terms of rich text prompt design. Moreover, manually crafting keypoint-specific prompts is a cumbersome task. As depicted in Figure~\ref{fig:intro}, a solitary textual description often fails to encapsulate all the nuances of a specific keypoint, such as its color, position, and shape. For example, the ``left shoulder" keypoint can be described from both spatial and color perspectives as ``Left shoulder is near the right shoulder" and ``The color of the left shoulder is light salmon", respectively. From a textual standpoint, multiple descriptions provide diverse perspectives, enabling pose estimation to grasp high-level semantics and pinpoint keypoints accurately based on one or more of these cues. On the visual front, the intricacies of wild scenes and the presence of multiple species introduce uncertain statistical shifts compared to the training domain \citep{jiang2022animal,lu2022few,wang2024sql}. Given this ``uncertainty" in potential data variations, synthesizing novel feature statistics variants to model diverse representations can bolster the network's robustness against unseen categories. To address these challenges, this paper introduces a novel probabilistic method that enhances network generalization to handle unseen categories by adeptly modeling keypoint-adaptive prompts.

\begin{figure*}[htbp]
  \centering
  \includegraphics[trim={0mm 0mm 0mm 0mm},clip,width=\textwidth]{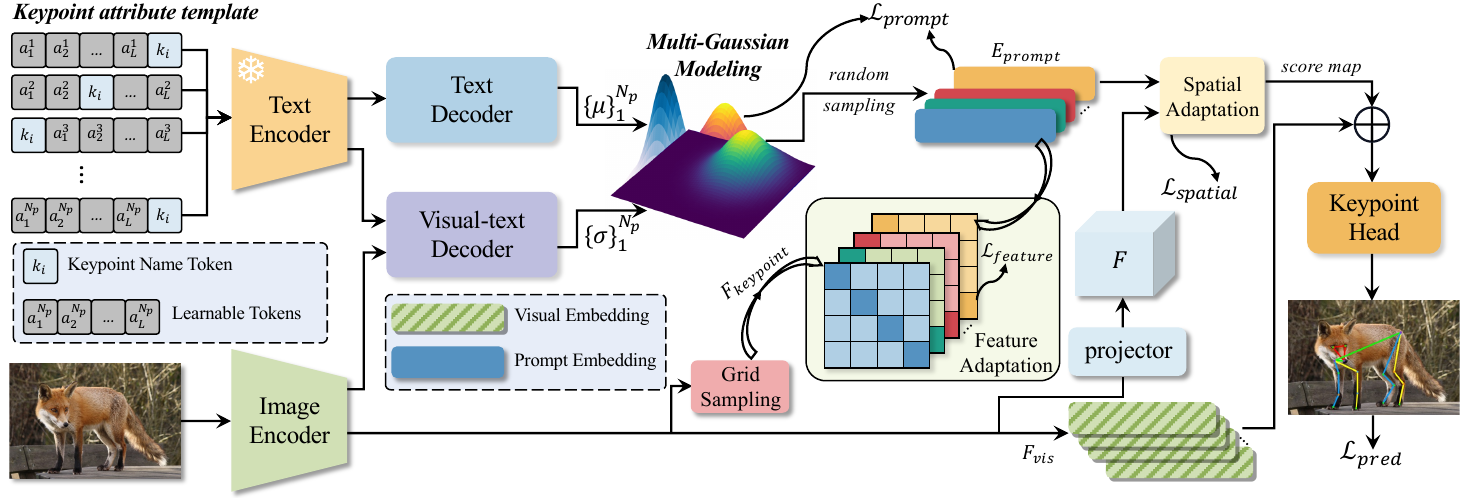}
  \hfill
  \vspace{-10pt}
  \caption{Overall framework of PPAP. Firstly, we create $N_p$ keypoint attribute templates and generate distinctive embeddings for these attributes using a text encoder. Each keypoint prompt embedding is represented probabilistically as a multivariate Gaussian distribution, with its mean derived from the text decoder and variance from the visual-text decoder. Subsequently, we sample these keypoint prompt representations from the distribution and perform cross-modal fusion during the spatial adaptation stage to capture the spatial relationship between textual and visual content.}
  \vspace{-15pt}
  \label{fig:framework}
\end{figure*}

Our approach leverages multiple learnable text prompts to enhance flexibility and robustness in text prompt initialization, moving beyond a single template. By strategically inserting keypoint names within templates, we enable flexible sentence construction akin to natural language. We hypothesize that prompt statistics, incorporating uncertainties, follow a multivariate Gaussian distribution, with the center defined by the original prompt statistic and the range indicating data variation across animal categories. We estimate this range using mini-batch statistics in a parametric learning framework and randomly sample variants from the estimated distribution to model diverse textual descriptions, as illustrated in Figure~\ref{fig:framework}. To capture diverse and informative keypoint attributes, we introduce a novel diversity loss function that ensures distinctiveness across prompt distributions.
In summary, our contributions are as follows:
\begin{itemize}
\item We propose a novel Probabilistic Prompt Distribution Learning framework for APE to learn the probabilistic representation of specific keypoint descriptions.
\item We introduce multiple prompt attributes and design a diversity loss to prevent the learned attributes from being identical. Additionally, we explore three different cross-modal fusion strategies to achieve spatial adaptation.
\item We demonstrate the effectiveness of the proposed probabilistic-based approach through extensive experiments on multi-species animal pose estimation.
\end{itemize}

\section{Related Works}
\label{sec:relatedwork}

%-------------------------------------------------------------------------
\subsection{Animal Pose Estimation}
Typical vision-only APE methods~\citep{cao2019cross,mathis2018deeplabcut,yu2021apk,rao2022kitpose,ng2022animal,xu2025learning}, with most directly adapting human pose estimation methods~\citep{xiao2018simple,sun2019deep,yuan2021hrformer,xu2022vitpose,liu2023group,yang2023effective}, such as DeepLabCut~\citep{lauer2022multi,mathis2018deeplabcut}, AP-10K~\citep{yu2021apk,sun2019deep} and ViTPose~\citep{xu2022vitpose,xu2023vitpose++} to animal counterpart.
These methods leverage powerful visual backbones for feature extraction but fail to handle rare or even unseen categories explicitly.
The relatively limited animal data volume compared to its human counterpart increases the risk of overfitting in complex networks~\citep{jiang2022animal}.
To cope with the limited labelled animal data volume, direct attempts~\citep{mu2020learning,li2021synthetic} learn prior-aware knowledge from synthetic animal datasets, and generate pseudo labels for unlabelled data.
In terms of refining the pseudo labels, ScarceNet~\citep{li2023scarcenet} and FreeNet\citep{zeng2024towards} propose the novel semi-supervised learning methods that improves the reliability of pseudo label generation, effectively mitigating the data scarcity problem.
Following the principles of few-shot learning, Xu et al.~\citep{xu2022pose} proposed a novel Category-Agnostic Pose Estimation (CAPE) paradigm, which treats keypoint prediction as the keypoint matching problem, enabling the prediction of poses on unseen categories.
% Following the seminal work, many variants~\citep{ren2024dynamic,nguyen2024escape} have followed this paradigm and proposed effective methods.
% However, the ability of CAPE-style methods to achieve category-agnostic pose estimation due to the introduction of category-specific prior knowledge during inference.
This is accomplished by offering a few support sets from the same category as the test image to guide the learning process.
CAPE~\citep{xu2022pose,ren2024dynamic,nguyen2024escape} introduces an additional computational burden and necessitates prior animal categories, making the methods less practical.
To eliminate the reliance on category-specific prior, pre-trained VLMs have been applied in the APE.
CLAMP~\citep{zhang2023clamp} is the first text-guided APE method, aligning text-visual features at both the feature and spatial levels, thereby enabling the textual prompt to effectively align the visual representation.
X-Pose~\citep{yang2025xpose} proposes an end-to-end pose estimation framework that leverages both visual and textual prompts to support keypoint prediction for arbitrary targets.
In view of this, we focus on designing keypoint prompts that offer enriched keypoint representations, facilitating accurate pose estimation.
\subsection{Deterministic Prompt Learning}
Prompt learning~\citep{zhou2022learning,zhou2022conditional,rao2022denseclip,khattak2023maple,khattak2023self} aims to adapt foundation models to downstream tasks by efficiently optimizing prompts.
CoOp~\citep{zhou2022learning} represents a seminal application of prompt learning in VLMs, transforming hard-coded discrete prompt templates into learnable continuous prompts.
CoCoOp~\citep{zhou2022conditional}, as its variants, integrates image-conditioned knowledge into text prompts to enhance generalization to unseen classes.
Similarly, DenseCLIP~\citep{rao2022denseclip} employs a Transformer to facilitate interaction between visual context and textual prompts, producing context-aware prompts for improved performance.
% MaPLe~\citep{khattak2023maple} uses both visual and textual prompts and progressively models the stage-wise multi-modal relationship to improve generalization.
Building upon this, PromptSRC~\citep{khattak2023self} uses a set of hand-crafted prompts and introduces several regularization losses to prompt learning.
To further improve diversity, we specialize different keypoint prompts to guide spatial adaptation for different animal categories, thus aligning them with different attribute sets for each keypoint. 
\subsection{Probabilistic Prompt Learning}
% Given a few samples of training dataset, the point estimate of text feature given prompt is hard to capture unseen image features.
Compared to deterministic prompt learning, probabilistic prompt learning ~\citep{cho2024make,lu2022prompt,kwon2023probabilistic} is not tied to specific classes, making it more adaptable to cross-species challenges with prompt adjustments.
 % is better suited to address the multi-species problem.
APP~\citep{cho2024make} proposes a Bayesian-based prompt learning framework to enhance the flexibility and adaptability of input embeddings.
However, various descriptions are dispersed in the input space, resulting in difficulty in representing their distribution.
Several studies focus on probabilistic learning within the output embedding space.
ProDA~\citep{lu2022prompt} is the first probabilistic prompt model, allowing for straightforward modeling by a multivariate Gaussian distribution.
PPL ~\citep{kwon2023probabilistic} introduces attribute prompts to represent shared attributes across different categories by constructing a mixture of Gaussian distribution to address dense prediction problems.
Conversely, we prioritize probabilistic modeling for each individual prompt rather than constructing a single overarching distribution, as it provides us with nuanced contextual information better and shows greater generalization to unseen data.

%-------------------------------------------------------------------------

\section{Method}
\label{sec:method}
% In order to enhance generalization capability on multi-species animal data, we propose a probabilistic prompt learning method. As the probabilistic prompt learning framework is based on Contrastive Learning for Connecting Language and Animal Pose (CLAMP)~\citep{zhang2023clamp}, we start with a brief summary of CLAMP's core principles and then present our probabilstic prompt learning method.

To enhance the generalization capabilities for multi-species animal data, we propose a probabilistic prompt learning technique. This technique is grounded in the Contrastive Learning for Connecting Language and Animal Pose (CLAMP) framework \citep{zhang2023clamp}. We begin by providing a concise overview of CLAMP's fundamental principles before presenting our probabilistic prompt learning method.

\subsection{Preliminaries}
CLAMP is proposed based on the powerful model CLIP~\citep{radford2021clip}.
Given an image and its corresponding textual description, CLIP extracts textual and visual embeddings by processing the input through a text encoder and an image encoder, respectively.
To effectively leverage the textual description for APE, CLAMP uses a pose-specific template $p_n=\{T_l\}_{l=1}^L[KeyPoint]_n$ to represent $n$-th keypoint prompt, where $\{T_l\}_{l=1}^L$ is $L$ learnable prefix tokens, and $[Keypoint]_n$ denotes the $n$-th keypoint name token. Then prompt embedding $E_{prompt}\in\mathbb{R}^{K\times C_{emb}}$ is obtained by the prompt encoder.
To establish spatial connections between textual description and image features, CLAMP performs cross-modal adaptation at both spatial and feature levels.
In the spatial adaptation stage, the extracted prompt representations and image features are used to generate score map $S\in\mathbb{R}^{K\times H\times W}$, where $H$, $W$ and $K$ represent the height, width and the number of keypoints, respectively.
This $S$ is utilized to indicate the probability of keypoint occurrence at specific locations. Besides, the spatial objective $\mathcal{L}_{spatial}$ is calculated as follows:
\begin{equation}
    \mathcal{L}_{spatial}=MSE(Upsample(S),H_{target}),
\label{eq:spat-adap}
\end{equation}
where $MSE$ denotes mean squared error, $Upsample$ is the upsample operation, and $H_{target}$ is the target heatmap.
Meanwhile, in the feature adaptation stage, ground truths are performed to grid-sample local keypoint features $F_{keypoint}\in\mathbb{R}^{K\times C_{emb}}$ from image embeddings.
The pairwise similarity $M=F_{keypoint}\cdot E_{prompt}^T$ is utilized to facilitate feature alignment:
\begin{equation}
    \mathcal{L}_{feature}=\frac12(CE(M,M_{label})+CE(M^T,M_{label})),
\label{eq:feat-adap}
\end{equation}
where $CE$ denotes cross-entropy loss and $M_{label}$ is the diagonal matrix, used as matching target.
Building on these two adaptations, we introduce a set of learnable keypoint-agnostic prompts, modeling each prompt distribution with a Gaussian.
Finally, we explore three fusion strategies for spatial adaptation and design a novel pose estimation paradigm to handle the multi-species APE.
% We represent the prompt embeddings and keypoint embeddings as $E_{prompt}\in\mathbb{R}^{N\times C_{emb}}$ and $F_{keypoint}\in\mathbb{R}^{N\times C_{emb}}$, respectively.

\subsection{Diverse Prompt Construction}
\label{subsec:diversePM}
Relying solely on visual features and the keypoint name prompt proves to be insufficient to address the long-tail problem present in animal datasets \citep{yu2021apk,ng2022animal}.
As previous literature \citep{lu2022prompt,kwon2023probabilistic} indicated, multiple diverse prompts could offer enriched semantics to aid multi-modal models in recognizing visual patterns, thereby alleviating the adverse impact of long-tail challenge.
However, it is hard to explicitly identify ``what constitutes useful keypoint visual attributes for locating keypoints?".
To this end, we introduce multiple learnable textual attribute prompts for each keypoint to constitute the prompt templates.
% Besides, different attributes can be defined by different distributions.
Formally, as illustrated in Figure~\ref{fig:framework}, for the $i$-th keypoint token $k_i$, we create $N_p$ keypoint attribute templates for each keypoint by combining learnable attribute description tokens.
\begin{equation}
    p_i^t=\{a_1^t,a_2^t,\ldots,a_L^t\ |\ k_i\},\quad t=1,\ldots,N_p,
\end{equation}
where $\{a_1^t,a_2^t,\ldots,a_L^t\}$ comprises $L$ learnable attribute tokens, initialized randomly with ``X X ... X".
As shown in Figure~\ref{fig:intro}, we aim to simulate the locations of keypoint in natural language descriptions.
To this end, we introduce the \textbf{Generalized Keypoint Placement} (GKP) strategy, enabling flexible placement of keypoint names within the prompt template for enhanced naturalness and diversity. Unlike the insertion strategy used in ProDA~\citep{lu2022prompt}, which restricts keypoint placement to specific positions (\ie beginning, middle, or end), our approach permits random placement throughout the template, offering greater positional flexibility. During training, these tokens are optimized to better suit APE tasks, offering increased contextual flexibility. By inputting the textual attribute prompts into the text encoder $\theta(\cdot)$, we can obtain the encoded diverse prompt representations:
\begin{equation}
    \boldsymbol{P}_i=\{\theta(p_i^t)|_{t=1}^{N_p}\},\quad \tilde{\boldsymbol{P}}_i=\{\theta(\tilde{p}_i^t)|_{t=1}^{N_p}\},
\end{equation}
where $\boldsymbol{P}_i$ denotes the complete embedding for the $i$-th keypoint prompt. Meanwhile, $\tilde{p}_i^t$ represents attribute tokens excluding keypoint names, and $\tilde{\boldsymbol{P}}_i$ signifies the keypoint-agnostic attribute embedding.

\textbf{Multiple Attributes Diversity Loss}.
To make the keypoint-agnostic attribute representations capture distinct content, we introduce diversity among representations, and design a novel diversity loss to keep differentiation between them.
The specific formulation is as follows:
\begin{equation}
\small
    % \mathcal{L}_{\text{div}}=\frac{1}{N_p^2}\sum_{i=1}^{N_p}\sum_{j=1}^{N_p}\parallel \tilde{\boldsymbol{P}}_i\tilde{\boldsymbol{P}}_j^T-\llbracket i=j \rrbracket \parallel_2^2
    \mathcal{L}_{div}=\frac1K\sum_{i=1}^K\parallel \tilde{\boldsymbol{P}}_i\tilde{\boldsymbol{P}}_i^T-\mathbb{I}\parallel_2^2,\quad \mathbb{I}=\text{diag}(\underbrace{1, 1, \dots, 1}_{N_p}).
\end{equation}
% \todo{The loss penalizes the attribute embeddings to approximate the identity matrix $\mathbb{I}\in\mathbb{R}^{N_p\times N_p}$ via an $l2$-norm minimization.}
The diversity loss applies a squared $\ell_2$-norm penalty on attribute representations to foster greater dissimilarity among them, thereby amplifying attribute diversity. Subsequently, the resultant prompt representations encapsulate the abundant and varied textual semantics, providing enhanced contextual support for subsequent visual localization.

\subsection{Probabilistic Prompt Modeling}
Based on the observation~\citep{lu2022prompt}: the output embedding of the descriptions within a category are close to each other, allowing them to be modeled with a sample distribution, \ie Gaussian.
We propose a probabilistic prompt learning strategy to reformulate the prompt distribution.
As subsection~\ref{subsec:diversePM} discussed, we strive to maintain orthogonality among the attribute embeddings.
Thus, these embeddings are better represented as independent Gaussian distributions, each characterized by distinct means and variances.
For each $p_i$, we define a probabilistic distribution $\mathcal{G}(z_i^t|p_i^t)$ as a factorized Gaussian with its center vector $\mu_i^t$ and variance matrix $\sigma_i^t$, where $t$ is the $t$-th attribute for the $i$-th keypoint. This distribution is formulated as:
\begin{equation}
    \mathcal{G}(z_i|p_i)\sim\{\mathcal{N}(\boldsymbol{\mu}_i^t,\boldsymbol{\sigma}_i^t\mathbf{I})\}_{t=1}^{N_p}.
\end{equation}
% We obtain the mean of Gaussian distribution $\boldsymbol{\mu}_i^t$ through interactions among attribute embeddings, while the variances $\boldsymbol{\sigma}_i^t$ are introduced through interactions between visual and textual modalities. 
Specifically, we introduce a text decoder to obtain the mean of Gaussian distribution $\boldsymbol{\mu}_i^t$ and a visual-text decoder for the variances $\boldsymbol{\sigma}_i^t$, as illustrated in Figure~\ref{fig:framework}.
In the text decoder, we facilitate interaction among prompt embeddings by self-attention ($SA$) module, as well as a form of self-augmentation. The default number of the layer is set to 1.
% , modeling the relationships between embeddings and generating the representation centers $\boldsymbol{\mu}_i^t$:
\begin{equation}
    \boldsymbol{\mu}_i^t = MLP(LN(q_i^t))+SA(LN(q_i^t,k_i^t,v_i^t)),
\end{equation}
where the query $q_i^t$, key $k_i^t$, and value $v_i^t$ are all derived from prompt embedding $P_i$. $LN$ and $MLP$ denote layer norm and multilayer perceptron layer, respectively.
% We feed the prompt representations $\boldsymbol{P}_i$ into the typical transformer module, whose output is subsequently combined with the visual features $F$ into the cross-attention ($CA$) module to generate the new $\bar q_i^t$.
Meanwhile, in the visual-text decoder
Here, the key $v_k$ and value $v_\text{v}$ vectors are both obtained from the $F$.
The variance matrix of each attribute representation is computed as:
\begin{equation}
    \boldsymbol{\sigma}_i^t=MLP(LN(q_i^t))+CA(LN(q_i^t,v_k,v_\text{v})),
\end{equation}
where CA stands for cross-attention module.
% Finally, we reformulate the prompt distribution for the keypoint as a multiple Gaussian model such that 
% The $\mathcal{G}(z_i|p_i)$ can be interpreted as a distribution of possible attribute representations that reflect visual-context knowledge from the input image.
Once the statistics of each keypoint description are obtained, the Gaussian distribution for probabilistic prompt statistics can be established.
By exploiting the Gaussian distribution, random sampling can generate various new feature statistics information with different means and variances. Therefore, we adopt random sampling to further exploit the probabilistic representations. $N_s$ prompt representations are sampled for $i$-th keypoint with $t$-th attribute $z_i^t=\{z_i^1,\dots,z_i^{N_s}\}\overset{\iid}{\sim}\mathcal{G}(z_i^t|p_i^t)$, and we interpret $z_i^t$ as independently enriched contextual description.
Here we use the re-parameterization trick~\citep{kingma2015variational} to make the sampling operation differentiable.
\begin{equation}
    \hat z_i^t=\mu(p_i^t)+\epsilon\cdot\sigma(p_i^t),\quad \epsilon\sim\mathcal{N}(\boldsymbol{0},\boldsymbol{I}),
\end{equation}
where $\mu(p_i^t)$, $\sigma(p_i^t)$ are mean and standard deviation of distribution $\mathcal{G}(z_i^t|p_i^t)$, and $\epsilon$ follows the standard normal distribution.
Finally, we obtain a set of diverse prompt representations $\hat z_i^t$ for each keypoint.
% We employ the additional KL divergence regularization between each keypoint distribution and Gaussian prior distribution $\mathcal{N}(0,\textit{I})$ to prevent the learned variance from collapsing to zero:
To prevent learned variances from collapsing to zero, we employ an additional KL divergence regularizer between prompt distributions and the Gaussian prior, forming the keypoint prompt loss:
\begin{equation}
\small
\mathcal{L}_{prompt}=\mathcal{L}_{div}+\frac1{N_p}\sum_t^{N_p}KL(\mathcal{G}(\hat z_i^t|p_i^t)\parallel\mathcal{N}(\boldsymbol{0},\boldsymbol{I})).
\end{equation}
\subsection{Cross-Modal Fusion}
Once various prompt embeddings have been obtained, the problem of ``how to align them with the visual features?" inevitably arises. Given these prompt representations, we explore three strategies for multi-modal information fusion: 1) \textbf{heuristic selection}; 2) \textbf{ensemble selection}; 3) \textbf{attention-based selection}.
Specifically, the \textbf{heuristic selection} strategy computes the similarity between the $N_s$ sampled score maps $S'\in\mathbb{R}^{N_s\times K\times H\times W}$ and the keypoints target $H_{target}\in\mathbb{R}^{K\times H\times W}$, assigning the most similar ones as the final score map $S$. Thus, the score map for the $i$-th keypoint is voted from all the sampling turns.
The \textbf{ensemble selection} strategy considers all the representations to contain informative context.
To prevent the loss of textual cues, we concat score maps and apply convolution operation ($Conv$) for signal modulation, ultimately generating the final score map.
\begin{equation}
    S=Conv(Concat(S')).
\end{equation}
For the \textbf{attention-based selection}, we introduce an additional learnable query to leverage the attention module for capturing relationships between the query embedding and the prompt representations.
We predefine the query embedding, \ie $q_i\in\mathbb{R}^{K\times (HW)}$ for learning the suitable prompt representation.
Meanwhile, the attention module takes the sampled prompts as key and value, which is formulated as:
\begin{equation}
    \begin{aligned}
        &S_i=Flatten(S'), \\
        &Q_i=q_iW_q,\ K_i=S_iW_k,\ V_i=S_iW_v, \\
        &S_i=Q_i+W_a(\text{softmax}(\frac{Q_iK_i^T}{\sqrt{HW}})V_i),
    \end{aligned}
\end{equation}
where $W_q, W_k, W_v$ are learnable linear projections, and $S_i$ represents $i$-th keypoint score map.
Subsequently, $S_i$ is reshaped back to 2D with the final score map $S=\{S_i\}_{i=1}^K$.
Unless otherwise specified, we choose two attention layers in our experiments as default.

\subsection{Training}
\textbf{Total objective}.
% Therefore, with the MSE regression loss, $\mathcal{L}_{\text{pred}}$ for the pose estimation task, the overall training loss can be written as a weighted summation of all loss functions:
For the pose estimation task, we employ the commonly used MSE loss $\mathcal{L}_{pred}$, and the total objective $\mathcal{L}_{total}$ is formulated as follows:
\begin{equation}
\footnotesize
\begin{aligned}
    &\mathcal{L}_{pred}=MSE(F_o,H_{target}),\quad F_o=HEAD(F_{
    vis}\oplus S), \\
    &\mathcal{L}_{total}= \mathcal{L}_{pred}+\mathcal{L}_{spatial}+\gamma\cdot\mathcal{L}_{feature} + \beta\cdot\mathcal{L}_{prompt},
\end{aligned}
\label{eq:objective}
\end{equation}
where $F_{vis}$ denotes the visual features output by image encoder, and $HEAD$ is the common keypoint prediction head~\citep{xiao2018simple}, which consists of multiple deconvolution and MLP layers.
The coefficient $\gamma$ and $\beta$ are hyper-parameters, which are discussed in detail in the ablation study section.
\section{Experiments}
\subsection{Setup}
\textbf{Datasets and evaluation metrics.}
(1) \textbf{AP-10K}~\citep{yu2021apk} is a large-scale benchmark for multi-species animal pose estimation and contains over 10K images collected and filtered from 23 animal families and 54 species with 17 key points. We employ the mean average precision (AP) as the primary evaluation metric. Average recall (AR) and APs at different OKS (Object Keypoint Similarity) thresholds are also reported, \ie AR, AP.5 and AP.75, as well as APs for different instance sizes, \ie AP$^{\text{M}}$ and AP$^{\text{L}}$. (2) \textbf{AnimalKingdom} \citep{ng2022animal} is a more challenging dataset, which includes over 33K samples with 23 keypoints across 850 species in 6 classes. Unlike AP-10K, AnimalKingdom contains a larger number of categories with more imbalanced and diverse data distribution, enhancing its difficulty. Besides, AnimalKingdom define three protocols (P1, P2, P3) for different tasks. P1-All involves training on the whole dataset for supervised learning. P2-LeaveKOut selects k sub-classes for training and performs few-shot learning on the remaining sub-classes. P3-Amphibians/Birds/Fishes/Mammals/Reptiles specifies training within each distinct animal class, enabling zero-shot learning on unseen classes. We report the Percentage of Correct Keypoint (PCK@$\alpha$) score for evaluation, which computes the percentage of correct keypoints whose distance to ground truth is within $\alpha$ times the longest bounding box side.
Besides, we adhere to the evaluation protocol in \citep{ng2022animal}, and set $\alpha=0.05$.

% -------------------------------------------------- AP10K Performance ----------------------------------------------
\begin{table*}[htbp]
\centering
\resizebox{\linewidth}{!}{
\tiny
\begin{threeparttable}
\begin{tabular}{l|c|l|ccccccc}
\toprule
% \multicolumn{2}{c|}{Methods}  
Methods & Dataset  & Backbone  & AP   & AP\scalebox{0.8}{$^{.50}$} & AP\scalebox{0.8}{$^{.75}$} & AP\scalebox{0.8}{$^{\text{M}}$} & AP\scalebox{0.8}{$^\text{L}$} & AR \\
\specialrule{0.1pt}{0.5pt}{0.5pt}
\multicolumn{9}{l}{\cellcolor{Gray!20} \textit{Foundation Models}} \\
ViTPose++~\citep{xu2023vitpose++} & Mixed Datasets(387K)    & MAE / ViT-Base    & 74.5 & 94.9    & 82.2      & 46.8     & 75.0   & 70.0 \\
ViTPose++~\citep{xu2023vitpose++} & Mixed Datasets(387K)    & MAE / ViT-Large     & 80.4  & 97.6  & 88.5  & 52.7  & 80.8  & - \\
SuperAnimal~\citep{ye2024superanimal}   & Quadruped-80K   & HRNet-W32 & 80.1 & - & - & - & - & - \\
X-Pose-V~\citep{yang2025xpose} & UniKPT(226K)  & Swin-Tiny    & 73.6  & 91.9  & 80.6  & 47.2   & 74.2 & -\\
X-Pose-V~\citep{yang2025xpose} & UniKPT(226K)  & Swin-Large    & 79.0  & 95.7  & 86.8  & 57.0  & 79.6  & - \\
\specialrule{0.1pt}{0.5pt}{0.5pt}
\multicolumn{9}{l}{\cellcolor{Gray!20} \textit{Expert Models}} \\
% \multirow{13}{*}{\rotatebox{90}{\textbf{Vision-Only}}}
SimpleBaseline~\citep{xiao2018simple} & AP-10K & ResNet-50  & 70.2  & 94.2  & 76.0  & 45.5  & 70.4  & 73.5 \\
% SimpleBaseline~\citep{xiao2018simple} &   AP-10K    & ResNet-50    & 70.9  & 94.6  & 76.8  & 44.8  & 71.2  & 74.1 \\
HRNet~\citep{sun2019deep}  & AP-10K     & HRNet-W32 & 73.8 & 95.8      & 80.3      & 52.3     & 74.2   & 76.9 \\
HRNet~\citep{sun2019deep}  & AP-10K    & HRNet-W48  & 74.4 & 95.9      & 80.7      & 58.9     & 74.8   & - \\
% DARK~\citep{zhang2020distribution}   & AP-10K      & HRNet-W32    & 74.6 & 95.9      & 80.8      & 62.6     & 74.9   & 77.8 \\
% TokenPose~\citep{li2021tokenpose}  & AP-10K   & B    & 69.2 & 93.5      & 75.9      & 54.7     & 69.7   & 72.7 \\
TokenPose~\citep{li2021tokenpose}  & AP-10K    & HRNet-W48   & 72.7 & 94.3      & 79.2      & 60.6     & 72.7   & 75.7 \\
TransPose~\citep{yang2021transpose}  & AP-10K    & HRNet-W48   & 74.2 & 95.7      & 80.8      & 56.2     & 74.6   & 77.0 \\
HRFormer~\citep{yuan2021hrformer}   & AP-10K    & HRFormer-B   & 74.5 & 95.9      & 81.6      & 55.8     & 74.9   & 77.6 \\
% KITPose~\citep{xu2025learning}  & AP-10K    & HRNet-W48 & 77.3  & 96.3  & 84.9  & 57.0  & 77.7  & 80.4 \\
% % &ViTPose-B  & AP-10K       & ViT-Base (MAE)    & 66.3 & 92.0      & 72.5      & 59.4     & 66.4   & 70.0 \\
% % &ViTPose-L  & AP-10K      & ViT-Large (MAE)    & 71.2 & 94.2      & 77.8      & 59.5     & 71.4   & 74.6 \\
\specialrule{0.1pt}{0.5pt}{0.5pt}
\multicolumn{9}{l}{\cellcolor{Gray!20} \textit{Prompt-based Models}} \\
% % \multirow{8}{*}{\rotatebox{90}{\textbf{Vision-Language}}}
CLAMP~\citep{zhang2023clamp} & AP-10K & CLIP / ResNet-50  & 72.9 & \textbf{95.4} & 79.4  & 43.2  & 73.2  & 76.3 \\
PPAP (Ours) & AP-10K  & CLIP / ResNet-50   & \textbf{73.4}  & 94.1  & \textbf{79.9}  & \textbf{57.1}  & \textbf{73.8}  & \textbf{76.5} \\
% \specialrule{0.1pt}{0.5pt}{0.5pt}
CLAMP~\citep{zhang2023clamp} & AP-10K  & CLIP / ViT-Base  & 74.3 & 95.8 & 81.4  & 47.6  & 74.9  & 77.5 \\
PPAP (Ours) & AP-10K  & CLIP / ViT-Base    & \textbf{75.4}  & \textbf{95.8}  & \textbf{83.0}  & \textbf{53.4}  & \textbf{75.9}  & \textbf{78.7} \\
% \specialrule{0.1pt}{0.5pt}{0.5pt}
CLAMP~\citep{zhang2023clamp} & AP-10K  & CLIP / ViT-Large  & 77.8 & 96.8 & 85.0  & 58.7  & 78.1  & 81.0 \\
PPAP (Ours) & AP-10K  & CLIP / ViT-Large    & \textbf{78.6}  & \textbf{97.1}  & \textbf{86.1}  & \textbf{59.3}  & \textbf{78.9}  & \textbf{81.7} \\
% \midrule 
\bottomrule
\end{tabular}
\end{threeparttable}
}
\vspace{-0.2em}
\caption{\textbf{Performance comparisons on AP-10K \textit{val} set}. The image resolution is $256\times256$, with an exception for ViTPose++, which utilizes a resolution of $256\times192$. Notably, the foundation models, such as ViTPose++, are trained on mixed datasets with extra corpora, which can be regarded as an upper bound for model performance.
}

\vspace{-0.6em}
\label{tab:ap10k}
\end{table*}
% -----------------------------------------------------------------------------------------------------------------

% ---------------------------------- Animal Kingdom Performance ------------------------------------
\begin{table*}[htbp]
\renewcommand{\arraystretch}{1.0} % set the space between rows
% \centering
% \setlength\tabcolsep{3.5pt} % set the column space
    \begin{subtable}[t]{0.65\textwidth}
     \setlength\tabcolsep{4pt}
        \raggedleft
        \footnotesize
        \begin{tabular}{l|c|cccccc}
        \toprule
        Methods & Backbone  & P3-Amph.  & P3-Birds   & P3-Fishes   & P3-Mam.    & P3-Rep.   & P1-All \\
        % \specialrule{0.1pt}{0.5pt}{0.5pt}
        % HRNet~\citep{sun2019deep}   & HRNet-W32 & 56.7 & 77.3   & 68.2 & 61.6   &  56.1 & 66.1 \\
        \specialrule{0.1pt}{0.5pt}{0.5pt}
        CLAMP~\citep{xiao2018simple}  & CLIP / ResNet-50  & 51.8   & 76.1  & 62.4  & 54.0  & 43.5  & 60.5 \\
        PPAP (Ours)    & CLIP / ResNet-50  & \textbf{51.9} & \textbf{76.5}  & \textbf{62.4}  & \textbf{55.5}  & \textbf{48.2} & \textbf{60.9} \\
        \specialrule{0.1pt}{0.5pt}{0.5pt}
        CLAMP~\citep{xiao2018simple}  & CLIP / ViT-Base  & 58.9   & 76.9  & 65.3  & 59.2  & 55.7  & 62.8 \\
        PPAP (Ours)    & CLIP / ViT-Base  & \textbf{60.9} & \textbf{78.0}  & \textbf{66.1}  & \textbf{61.2}  & \textbf{56.6} & \textbf{64.5} \\
        \bottomrule
        \end{tabular}%
        \label{tab:ak-1}
        \caption{Comparisons across various protocols on AnimalKingdom.}
    \end{subtable}
    \hfill
    \begin{subtable}[t]{0.3\textwidth}
    \setlength\tabcolsep{5pt}
        \footnotesize
        \raggedright
        \begin{tabular}{l|c|c}
            \toprule
            Methods & Backbone  & P1-All \\
            \specialrule{0.1pt}{0.5pt}{0.5pt}
            HRNet~\citep{sun2019deep}   & HRNet-W32 & 66.1 \\
            DARK~\citep{zhang2020distribution}    & HRNet-W32 & 66.6 \\ \specialrule{0.1pt}{0.5pt}{0.5pt}
            CLAMP~\citep{zhang2023clamp}   & ViT-Large & 65.6 \\
            \textbf{PPAP (Ours)}    & \textbf{ViT-Large} & \textbf{68.4} \\
        \bottomrule
        \end{tabular}
        \label{tab:ak-2}
        \caption{Comparisons with the SOTA methods.}
    \end{subtable}
\vspace{-0.8em}
\caption{\textbf{Performance comparisons on the AnimalKingdom \textit{test} set (PCK$@$0.05).}
The input resolution of all methods is $256\times256$.}
\label{tab:animalkingdom}
\vspace{-1em}
\end{table*}
% -----------------------------------------------------------------------------------------------

\textbf{Implementation details.}
We follow the widely-used two-stage pose estimation paradigm in our experiments.
The ground-truth bounding box annotations are utilized to crop the instance target, which is subsequently resized to a uniform shape.
We select the representative CNN-based model ResNet alongside the transformer-based model ViT as backbones, serving as feature extractors in our framework.
The text encoder after CLIP pre-training is adopted as our language models and initialized with the corresponding CLIP pre-trained weights.
For prompt learning, we configure the length of the learnable context tokens to match CLAMP, setting $L=8$ for fair comparisons.
% The keypoint head is a common detection head, consisting of multiple deconvolution and MLP layers in combination.

%During training, we set the input image size to $256\times256$ and undergo common data augmentation techniques, including random rotation, scaling, horizontal flipping, and half-body transformation. Each model is trained for a total of 210 epochs with a step-wise learning rate scheduler which decays by 10 at the 170th and 200th epoch, respectively. Our network is trained using the AdamW optimizer with a weight decay of 2.5$e$-5. We freeze the text encoder to conserve the pre-trained language knowledge. Furthermore, we conduct supervised learning and zero-shot learning to thoroughly evaluate the models' performance. For both supervised and zero-shot learning on AP-10K and AnimalKingdom, we train the models with a batch size of 64 and set 3$e$-4 as the initial learning rate. To facilitate comparative experiments, we set the number of attributes $N_p=2$. From these $N_p$ embeddings, we perform uniform sampling, with the turns $N_s$ equal to $N_p$.

During training, we use $256\times256$ input images and applied data augmentation including random rotation, scaling, flipping, and half-body transformation. All models are trained for 210 epochs with a step-wise learning rate scheduler that decreases by a factor of 10 at the 170th and 200th epochs. We used the AdamW optimizer with a weight decay of $2.5e-5$ and froze the text encoder to preserve pre-trained language knowledge. Supervised learning experiments were conducted to evaluate performance on both AP-10K and AnimalKingdom, with a batch size of 64, an initial learning rate of $3e-4$, and $N_p=N_s=2$ for attribute embeddings and sampling turns, respectively. %While zero-shot learning experiments were conducted on the more diverse and challenging AnimalKingdom dataset under the same training settings as supervised learning.

\subsection{Results and Analysis}
\subsubsection{Supervised Learning.}
Table~\ref{tab:ap10k} presents the results under the supervised learning setting on \textbf{AP-10K}~\citep{yu2021apk}, comparing our method with representative expert models for pose estimation. It can be observed that probabilistic prompt learning contributes to improved performance in multi-species APE under the prompt learning paradigm. With the basic ResNet-50 as the feature extractor, PPAP achieves a modest AP improvement of 0.5, which we attribute to the limited capacity of the feature extractor in capturing sufficient image features. When switching to the more powerful ViT-Base backbone, PPAP achieves 75.4 AP and outperforms CLAMP by 1.1 points. These observations indicate that our prompt learning approach offers a more effective solution to the challenges posed by significant data variance in multi-species Animal Pose Estimation (APE), as opposed to merely relying on the keypoint name prompt.

Furthermore, we conduct comparisons against expert pose estimation models, validating our approach's effectiveness. On the AP-10K validation set, our PPAP method significantly outperforms the SimpleBaseline by 3.2 AP when using the ResNet-50 backbone, compared to the state-of-the-art CNN-based expert models. Additionally, when equipped with the ViT-Base backbone, our approach achieves 1.6 and 1.0 AP gains over the highly representative pose estimation model HRNet, respectively. In particular, PPAP also demonstrates notable performance improvements of 2.7 and 1.2 AP, respectively, when compared to hybrid architectures such as TokenPose and TransPose. We also observe that these methods underperform on multi-species animal datasets compared to human ones, all falling below HRNet. This is due to their inability to manage large data variances, highlighting our approach's effectiveness in overcoming this challenge. Besides, PPAP surpasses HRFormer by 0.9 AP when compared to a plain transformer-based architecture.
% Notably, PPAP with a ViT-Large backbone achieves 78.6 AP, approaching the UniPose-V with Swin-Large backbone.
% rivals state-of-the-art foundation models.

It's worth mentioning that foundation models are trained across multiple datasets like Quadruped-80K~\citep{ye2024superanimal} and UniKPT-226K~\citep{yang2025xpose}, allowing them to gain broader knowledge, richer pose priors, and generally enhanced performance. Despite this, our PPAP, trained exclusively on AP-10K, matches foundation models with a score of 78.6 vs. 79.0. This indicates that our method's robustness stems not from extensive pose data but from its effective utilization of textual knowledge to tackle multi-species variability.

% -------------------------------- zero-shot results ----------------------------------
\begin{table}[t]
    \footnotesize
    \centering
        \begin{tabular}{@{}c|l|cc|l@{}}
        \toprule
        \multicolumn{2}{c|}{Method} & Train  & Test  & PCK$@0.05$ \\
        \midrule
        \multirow{6}{*}{\textbf{ResNet-50}} &CLAMP & Fishes & Birds  & 9.5 \\
        &PPAP & Fishes  & Birds  & \textbf{9.8} (\red{$\uparrow 0.3$}) \\
        \cmidrule{2-5}
        &CLAMP  & Mammals   & Amphibians  & 6.8 \\
        &PPAP   & Mammals   & Amphibians  & \textbf{9.8} (\red{$\uparrow 3.0$}) \\
        \cmidrule{2-5}
        &CLAMP  & Reptiles   & Amphibians  & 19.2 \\
        &PPAP   & Reptiles   & Amphibians  & \textbf{19.3} \\
        \specialrule{0.1pt}{0.5pt}{0.5pt}
        \multirow{6}{*}{\textbf{ViT-Base}}&CLAMP & Fishes  & Birds & 0.1 \\
        & PPAP  &  Fishes   & Birds  & \textbf{12.0} (\red{$\uparrow 11.9$}) \\
        \cmidrule{2-5}
        & CLAMP & Mammals   & Amphibians    & 12.3 \\
        & PPAP  & Mammals   & Amphibians    & \textbf{16.5} (\red{$\uparrow 4.2$})  \\
        \cmidrule{2-5}
        & CLAMP & Reptiles  & Amphibians    & 20.8 \\
        & PPAP  & Reptiles  & Amphibians    & \textbf{21.8} (\red{$\uparrow 1.0$})  \\
        \bottomrule
        \end{tabular}
    \vspace{-5pt}
    \caption{\textbf{Zero-shot performance} comparison on AnimalKingdom.}
    \label{tab:zs-ak}
    \vspace{-5pt}
\end{table}
\begin{table}[t]
  \centering
  \resizebox{0.9\columnwidth}{!}{
  {\small	
    \begin{tabular}{p{1.5cm}p{1.5cm}p{1.5cm}|p{1cm}p{1cm}}
    \toprule
    \multicolumn{1}{c}{$\mathcal{L}_{\text{div}}$} & \multicolumn{1}{c}{$\mathcal{L}_{\text{KL}}$} & \multicolumn{1}{c|}{$\text{GKP}$} & \multicolumn{1}{c}{AP} & \multicolumn{1}{c}{$\Delta$} \\
    \midrule
          &       &       & \multicolumn{1}{c}{74.3}    & \multicolumn{1}{c}{0} \\
    \multicolumn{1}{c}{\checkmark}  &       &       & \multicolumn{1}{c}{74.9}  & \multicolumn{1}{c}{+0.6} \\
          & \multicolumn{1}{c}{\checkmark}  &  & \multicolumn{1}{c}{74.9}  & \multicolumn{1}{c}{+0.6} \\
    \multicolumn{1}{c}{\checkmark}  &  \multicolumn{1}{c}{\checkmark}     &   & \multicolumn{1}{c}{75.2}  & \multicolumn{1}{c}{+0.9} \\
    \multicolumn{1}{c}{\checkmark}  & \multicolumn{1}{c}{\checkmark}  & \multicolumn{1}{c|}{\checkmark}  & \multicolumn{1}{c}{\textbf{75.4}}  & \multicolumn{1}{c}{+1.1} \\
    \bottomrule
    \end{tabular}}}
    \vspace{-5pt}
    \caption{\textbf{Ablation studies} on components.}
    \vspace{-15pt}
  \label{tab:abl_loss}%
\end{table}%
% -------------------------------------------------------------------------------------

In addition to AP-10K, we further evaluate the effectiveness of PPAP on the more challenging dataset, \textbf{AnimalKingdom}~\citep{ng2022animal}.
The training setting is the same as the previous section.
Table~\ref{tab:animalkingdom} reports the performance of prompt-based methods across various protocols on the AnimalKingdom dataset.
We observe that the PPAP method consistently achieves superior performance across all protocols.
By constructing diverse prompts and employing probabilistic modeling, our PPAP gets gains of 0.4 and 1.7 PCK values with ResNet-50 and ViT-Base, respectively.
Notably, the considerable performance gains on specific classes underscore the \textit{adaptability} of our method.
For example, PPAP with ResNet-50 achieves a 4.7 PCK improvement in reptiles, while PPAP with ViT-Base yields a 3.0 PCK increase in mammals.
Compared to the existing SOTA methods, our PPAP with ViT-Large outperforms them by 2.0 AP gains.
Furthermore, we argue that incorporating the plug-and-play DARK~\citep{zhang2020distribution} leads to an additional boost in performance. Therefore, by training exclusively on the single pose data, our method achieves a new state-of-the-art performance with the ViT-Large backbone on both multi-species animal pose benchmarks. 

\begin{figure}[t]
  \centering
    % \begin{subfigure}[b]{0.88\linewidth}
    % \centering
    \includegraphics[width=\linewidth]{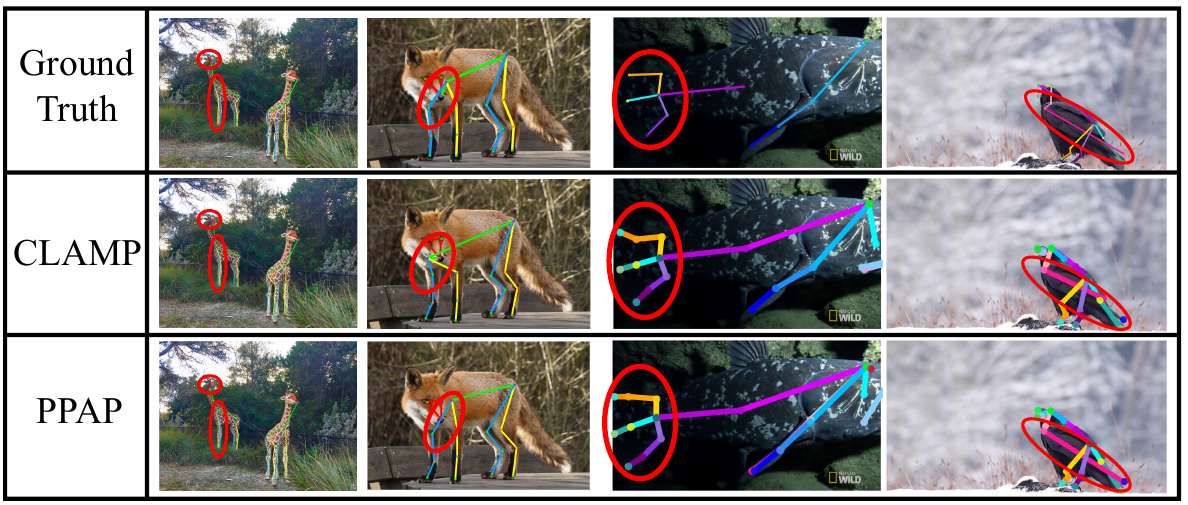}
    % \caption{Qualitative results for multi-species APE on AP-10K and AnimalKingdom based on CLAMP and our proposed PPAP. The first two columns show results on the AP-10K \textit{val} set, while the last two columns provide examples on the AnimalKingdom \textit{test} set.}
    % \end{subfigure}
    % \vfill
    % \begin{subfigure}[b]{0.88\linewidth}
    % \centering
    % \includegraphics[width=\textwidth]{figures/vis_emb.pdf}
    % \caption{T-SNE visualization of prompt representations on AP-10K dataset.}
    % \end{subfigure}
   \vspace{-10pt}
   \caption{Qualitative visualization of several examples with CLAMP and our proposed PPAP. }
   \label{fig:vis-quality}
   \vspace{-15pt}
\end{figure}

Besides, to intuitively highlight the superiority of our approach, we present qualitative comparisons with CLAMP in Figure~\ref{fig:vis-quality}.
Since AP-10K consists of mammals, we specifically select fishes and birds from AnimalKingdom for diversity.
Differences in predicted locations are highlighted in red circles, clearly illustrating that our method consistently achieves more precise localization across these three animal classes.
\begin{table}[t]
  \centering
  \renewcommand{\arraystretch}{1.2}
  \resizebox{1.0\columnwidth}{!}{
    \begin{tabular}{p{3.5cm}p{1cm}p{1cm}p{1cm}p{1cm}}
    \specialrule{0.1pt}{0.5pt}{0.5pt}
    \rowcolor[rgb]{ .949,  .949,  .949}\multicolumn{5}{l}{\textbf{Number of attributes} $(\gamma=5e\text{-4},\beta=1e\text{-5}, N_s=N_p$)} \\
    Parameter $N_p$ & $1$   & $2$   & $3$   & $4$ \\
    %\midrule
    AP  & 74.8     &\textbf{75.4}     & 75.1     & 75.0 \\
    \specialrule{0.1pt}{0.5pt}{0.5pt}
    \rowcolor[rgb]{ .949,  .949,  .949}\multicolumn{5}{l}{\textbf{Objective hyper-parameter} $(N_s=N_p=2)$} \\
    Parameter $\beta$ ($\gamma$=5$e$-4) & 1$e$-4 & 5$e$-5 & 1$e$-5 & 1$e$-6 \\
    %\midrule
    AP    & 74.5    & 74.8    & \textbf{75.4}  & 74.9 \\ \arrayrulecolor{gray}\hline  %\hdashline
    \arrayrulecolor{black}
    Parameter $\gamma$ ($\beta$=1$e$-5) & 5$e$-3 & 5$e$-4 & 5$e$-5 & 5$e$-6 \\
    %\midrule
    AP    & 73.5    & \textbf{75.4}    & 74.7  & 75.1 \\
    \specialrule{0.1pt}{0.5pt}{0.5pt}
    \rowcolor[rgb]{ .949,  .949,  .949}\multicolumn{5}{l}{\textbf{Number of sampled turns} $(\gamma=5e\text{-4},\beta=1e\text{-5}, N_p=2)$} \\
    Parameter $N_s$ & $2$   & $4$  & $6$  & $8$ \\
    %\midrule
    AP  & \textbf{75.4} & 74.7  & 74.5   & 74.6 \\
    \bottomrule
    \end{tabular}}
    \vspace{-5pt}
  \caption{\textbf{Ablation studies} on hyper-parameters.
  The number of attributes $N_p$, objective hyper-parameters $\gamma, \beta$, and the number of sampled representations $N_s$.}\vspace{-8pt}
  \label{tab:abbl_param}%
\end{table}%
\begin{table}[t!]
  \centering
  \resizebox{0.8\columnwidth}{!}{
    \small
    \begin{tabular}{p{3.5cm}|p{0.6cm}p{0.6cm}p{0.6cm}}
    \toprule
    Strategy & AP & AP$^{.50}$ & AP$^{.75}$ \\
    \midrule
    heuristic selection &  75.1  & 95.6   & 82.1  \\
    \textbf{ensemble selection}  &  \textbf{75.4} &  \textbf{95.8}   & \textbf{83.0} \\
    attention-based selection & 74.6 & 95.7  & 82.3 \\
    \bottomrule
    \end{tabular}}
    \vspace{-5pt}
    \caption{\textbf{Ablation studies} on cross-modal fusion strategies.}
    \vspace{-15pt}
  \label{tab:abl-sel}%
\end{table}%
\subsubsection{Zero-Shot Learning.}
Given that the AnimalKingdom dataset includes a broader range of classes and presents a more challenging benchmark, we further conduct zero-shot learning experiments on it to evaluate the model's generalization ability on unseen animal species.
Since the AnimalKingdom is sourced from wildlife data, it inevitably exhibits a long-tail distribution across various animal species~\citep{ng2022animal}.
We select three types of zero-shot settings to compare with the prompt-based method CLAMP~\citep{zhang2023clamp}.
The experiment settings include similar classes (``Reptiles vs. Amphibians"), typical classes (``Mammals vs. Amphibians"), and highly divergent classes (``Fishes vs. Birds"), thereby covering a range of taxonomic variations to fully reflect the generalization of the PPAP for dealing with unseen species.
The results are shown in Table~\ref{tab:zs-ak}, which reflects that our PPAP achieves better performance for all three settings.
While PPAP achieves only a modest performance gain with ResNet-50, it demonstrates a significantly greater improvement when paired with the more powerful ViT-Base backbone.
Interestingly, when trained on Fish and tested on Birds, CLAMP fails to perform pose estimation across such extreme inter-classes variations, resulting in nearly zero PCK value, while PPAP achieves a substantial improvement of 11.9 gains.
Furthermore, our method demonstrates additional performance gains of 4.2 and 1.0 in the other two settings, respectively.
Such observation clearly illustrates the generalization capability of our probability-based modeling approach on multi-species APE, effectively mitigating challenges associated with large data variance and imbalanced data distribution.

\subsection{Ablation Study}
%In this section, we perform an extensive ablation study of the proposed method on AP-10K with a ViT-Base backbone network. The effects of different components and hyper-parameters of the proposed method are analyzed below. Meanwhile, we also explore the different choices of cross-modal fusion strategies for obtaining the score map.

In this section, we conduct a comprehensive ablation study of our proposed method on the AP-10K dataset using ViT-Base backbone. We analyze the impact of various components and hyper-parameters, and also investigate different cross-modal fusion strategies for generating the score map.

\textbf{Effects of Different Components.}
%We conduct experiments to observe the effect of different loss functions and the Generalized Keypoint Placement (GPK) strategy. To this end, we train different models using the objective functions, and the results are reported in Table~\ref{tab:abl_loss}. As we can observe after constructing diverse prompts and applying constraint $\mathcal{L}_{\text{div}}$, the model achieves a 74.9 AP score and outperforms the baseline by 0.6 points. A similar observation is evident with the constraint $\mathcal{L}_{\text{KL}}$ applied Gaussian distribution reformulation, where it also yields a 0.6 improvement. Together with the two constraints, the model achieves 75.2 AP, surpassing the baseline by a large margin. Finally, by employing the GKP strategy, more generalized prompts are generated, facilitating further improvements.
We conducted experiments to evaluate the impact of various loss functions and the Generalized Keypoint Placement (GKP) strategy. Table~\ref{tab:abl_loss} summarizes the results. By constructing diverse prompts and applying the constraint $\mathcal{L}_{\text{div}}$, the model achieved a 74.9 AP score, outperforming the baseline by 0.6 points. Similarly, applying the constraint $\mathcal{L}_{\text{KL}}$ with Gaussian distribution reformulation also improved the AP score by 0.6 points. Combining both constraints, the model reached 75.2 AP, significantly surpassing the baseline. Additionally, using the GKP strategy generated more generalized prompts, leading to further improvements.

\textbf{Effects of Hyper-parameter.}
The hyper-parameters of the number of attributes $N_p$, objective function coefficient $\gamma, \beta$ in Eq.~\ref{eq:objective} and the sampling turns $N_s$ are to trade off the model performance. Firstly, we compare performance with respect to different attribute numbers $N_p$. As shown in the first block of Table~\ref{tab:abbl_param}, two prompt templates are sufficient to represent different image attributes and achieve optimal performance. As the number of prompt templates increases, however, performance declines, suggesting that excessive template quantity can lead to information redundancy, thereby impairing model efficacy. Then, we adjust the weighting ratio of the $\mathcal{L}_{feature}$ and $\mathcal{L}_{prompt}$, and observe that $\gamma = 5e-4$, $\beta = 1e-5$ are more suitable.  It suggests excessive weighting leads each prompt representation to converge toward a standard normal distribution, thereby diminishing their distinctiveness. Conversely, overly small weights result in excessive variance in the distribution of each embedding, which increases uncertainty. To achieve a balance, we choose $\beta = 1e-5$ to get the optimal result. In our approach, we employ reparameterization techniques to ensure differentiability, while also utilizing sampling to enhance the diversity of the text prompts. As we can see in the last block of Table~\ref{tab:abbl_param}, the results are not sensitive to the sampling turns and the accuracy reaches the best performance when setting $N_s$ and $N_p$ to be equal.

\textbf{Choices of Cross-modal Fusion.}
In our experiments, we explore three different fusion strategies, with results summarized in Table~\ref{tab:abl-sel}.
As we can see, no matter which strategy is employed, the performances are consistently higher than the baseline method.
Selecting the representation most similar to the target can achieve competitive performance.
However, as similarity computation relies on labeled data unavailable during testing, we instead average the sampled representations.
This discrepancy between training and testing leads to suboptimal performance.
Directly ensembling all representations followed by the conv modulation yields the best performance by avoiding information loss.
Besides, the results show that the attention-based strategy has worse performances, which may be attributed to the randomly initialized queries to effectively learn the suitable prompt representations.

% \subsection{Visualization and Analysis}
\section{Conclusion}
\label{sec-conclusion}
In this paper, we propose a probabilistic prompt distribution learning method to adapt pre-trained vision-language models for the multi-species APE task. With the diverse keypoint prompt construction, probabilistic prompt modeling and cross-modal fusion strategies, PPAP uses text guidance in a more comprehensive and effective manner, presenting a novel solution to the multi-species problem, \ie large data variance and imbalanced data distribution.
% Extensive experiments on the representative multi-species benchmarks, AP-10K and AnimalKingdom, demonstrate significant performance improvements with PPAP in both supervised learning and zero-shot learning settings.
PPAP reflects superiority and generalization for multi-species APE on the representative multi-species animal pose benchmarks including AP-10K and AnimalKingdom.

% \noindent\textbf{Acknowledgements}. This work was supported in part by the National Key Research and Development Project under Grant 2023YFC3806000, in part by the National Natural Science Foundation of China under Grant 62406226, and in part sponsored by Shanghai Sailing Program under Grant 24YF2748700.
\section*{Acknowledgements}
\label{sec-ack}
% \noindent\textbf{Acknowledgements}.
This work was supported in part by the National Key Research and Development Project under Grant 2023YFC3806000, in part by the National Natural Science Foundation of China under Grant 62406226 and 61936014, in part sponsored by Shanghai Sailing Program under Grant 24YF2748700, and in part sponsored by Tongji University Independent Original Cultivation Project under Grant 20230106.
{
    \small
    \bibliographystyle{ieeenat_fullname}
    \bibliography{main}

\begin{thebibliography}{44}
\providecommand{\natexlab}[1]{#1}
\providecommand{\url}[1]{\texttt{#1}}
\expandafter\ifx\csname urlstyle\endcsname\relax
  \providecommand{\doi}[1]{doi: #1}\else
  \providecommand{\doi}{doi: \begingroup \urlstyle{rm}\Url}\fi

\bibitem[Cao et~al.(2019)Cao, Tang, Fang, Shen, Lu, and Tai]{cao2019cross}
Jinkun Cao, Hongyang Tang, Hao-Shu Fang, Xiaoyong Shen, Cewu Lu, and Yu-Wing Tai.
\newblock Cross-domain adaptation for animal pose estimation.
\newblock In \emph{ICCV}, pages 9498--9507, 2019.

\bibitem[Cho et~al.(2024)Cho, Bae, Shin, Youn, Joo, and Moon]{cho2024make}
Youngjae Cho, HeeSun Bae, Seungjae Shin, Yeo~Dong Youn, Weonyoung Joo, and Il-Chul Moon.
\newblock Make prompts adaptable: Bayesian modeling for vision-language prompt learning with data-dependent prior.
\newblock In \emph{AAAI}, pages 11552--11560, 2024.

\bibitem[Davies(2015)]{davies2015keep}
Kay Davies.
\newblock Keep the directive that protects research animals.
\newblock \emph{Nature}, 521\penalty0 (7550):\penalty0 7--7, 2015.

\bibitem[Jiang et~al.(2022)Jiang, Lee, Teotia, and Ostadabbas]{jiang2022animal}
Le Jiang, Caleb Lee, Divyang Teotia, and Sarah Ostadabbas.
\newblock Animal pose estimation: A closer look at the state-of-the-art, existing gaps and opportunities.
\newblock \emph{Computer Vision and Image Understanding}, 222:\penalty0 103483, 2022.

\bibitem[Khattak et~al.(2023{\natexlab{a}})Khattak, Rasheed, Maaz, Khan, and Khan]{khattak2023maple}
Muhammad~Uzair Khattak, Hanoona Rasheed, Muhammad Maaz, Salman Khan, and Fahad~Shahbaz Khan.
\newblock Maple: Multi-modal prompt learning.
\newblock In \emph{CVPR}, pages 19113--19122, 2023{\natexlab{a}}.

\bibitem[Khattak et~al.(2023{\natexlab{b}})Khattak, Wasim, Naseer, Khan, Yang, and Khan]{khattak2023self}
Muhammad~Uzair Khattak, Syed~Talal Wasim, Muzammal Naseer, Salman Khan, Ming-Hsuan Yang, and Fahad~Shahbaz Khan.
\newblock Self-regulating prompts: Foundational model adaptation without forgetting.
\newblock In \emph{ICCV}, pages 15190--15200, 2023{\natexlab{b}}.

\bibitem[Kingma et~al.(2015)Kingma, Salimans, and Welling]{kingma2015variational}
Durk~P Kingma, Tim Salimans, and Max Welling.
\newblock Variational dropout and the local reparameterization trick.
\newblock In \emph{NeurIPS}, 2015.

\bibitem[Kwon et~al.(2023)Kwon, Song, Jeong, Kim, Jang, and Sohn]{kwon2023probabilistic}
Hyeongjun Kwon, Taeyong Song, Somi Jeong, Jin Kim, Jinhyun Jang, and Kwanghoon Sohn.
\newblock Probabilistic prompt learning for dense prediction.
\newblock In \emph{CVPR}, pages 6768--6777, 2023.

\bibitem[Lauer et~al.(2022)Lauer, Zhou, Ye, Menegas, Schneider, Nath, Rahman, Di~Santo, Soberanes, Feng, et~al.]{lauer2022multi}
Jessy Lauer, Mu Zhou, Shaokai Ye, William Menegas, Steffen Schneider, Tanmay Nath, Mohammed~Mostafizur Rahman, Valentina Di~Santo, Daniel Soberanes, Guoping Feng, et~al.
\newblock Multi-animal pose estimation, identification and tracking with deeplabcut.
\newblock \emph{Nature Methods}, 19\penalty0 (4):\penalty0 496--504, 2022.

\bibitem[Li and Lee(2021)]{li2021synthetic}
Chen Li and Gim~Hee Lee.
\newblock From synthetic to real: Unsupervised domain adaptation for animal pose estimation.
\newblock In \emph{CVPR}, pages 1482--1491, 2021.

\bibitem[Li and Lee(2023)]{li2023scarcenet}
Chen Li and Gim~Hee Lee.
\newblock Scarcenet: Animal pose estimation with scarce annotations.
\newblock In \emph{CVPR}, pages 17174--17183, 2023.

\bibitem[Li et~al.(2021)Li, Zhang, Wang, Yang, Yang, Xia, and Zhou]{li2021tokenpose}
Yanjie Li, Shoukui Zhang, Zhicheng Wang, Sen Yang, Wankou Yang, Shu-Tao Xia, and Erjin Zhou.
\newblock Tokenpose: Learning keypoint tokens for human pose estimation.
\newblock In \emph{ICCV}, pages 11313--11322, 2021.

\bibitem[Liu et~al.(2023)Liu, Chen, Tan, Liu, Wang, Su, Li, Yao, Han, Ding, et~al.]{liu2023group}
Huan Liu, Qiang Chen, Zichang Tan, Jiang-Jiang Liu, Jian Wang, Xiangbo Su, Xiaolong Li, Kun Yao, Junyu Han, Errui Ding, et~al.
\newblock Group pose: A simple baseline for end-to-end multi-person pose estimation.
\newblock In \emph{ICCV}, pages 15029--15038, 2023.

\bibitem[Lu and Koniusz(2022)]{lu2022few}
Changsheng Lu and Piotr Koniusz.
\newblock Few-shot keypoint detection with uncertainty learning for unseen species.
\newblock In \emph{CVPR}, pages 19416--19426, 2022.

\bibitem[Lu et~al.(2022)Lu, Liu, Zhang, Liu, and Tian]{lu2022prompt}
Yuning Lu, Jianzhuang Liu, Yonggang Zhang, Yajing Liu, and Xinmei Tian.
\newblock Prompt distribution learning.
\newblock In \emph{CVPR}, pages 5206--5215, 2022.

\bibitem[Mathis et~al.(2018)Mathis, Mamidanna, Cury, Abe, Murthy, Mathis, and Bethge]{mathis2018deeplabcut}
Alexander Mathis, Pranav Mamidanna, Kevin~M Cury, Taiga Abe, Venkatesh~N Murthy, Mackenzie~Weygandt Mathis, and Matthias Bethge.
\newblock Deeplabcut: markerless pose estimation of user-defined body parts with deep learning.
\newblock \emph{Nature Neuroscience}, 21\penalty0 (9):\penalty0 1281--1289, 2018.

\bibitem[Mu et~al.(2020)Mu, Qiu, Hager, and Yuille]{mu2020learning}
Jiteng Mu, Weichao Qiu, Gregory~D Hager, and Alan~L Yuille.
\newblock Learning from synthetic animals.
\newblock In \emph{CVPR}, pages 12386--12395, 2020.

\bibitem[Ng et~al.(2022)Ng, Ong, Zheng, Ni, Yeo, and Liu]{ng2022animal}
Xun~Long Ng, Kian~Eng Ong, Qichen Zheng, Yun Ni, Si~Yong Yeo, and Jun Liu.
\newblock Animal kingdom: A large and diverse dataset for animal behavior understanding.
\newblock In \emph{CVPR}, pages 19023--19034, 2022.

\bibitem[Nguyen et~al.(2024)Nguyen, Li, and Lee]{nguyen2024escape}
Khoi~Duc Nguyen, Chen Li, and Gim~Hee Lee.
\newblock Escape: Encoding super-keypoints for category-agnostic pose estimation.
\newblock In \emph{CVPR}, pages 23491--23500, 2024.

\bibitem[Pereira et~al.(2019)Pereira, Aldarondo, Willmore, Kislin, Wang, Murthy, and Shaevitz]{pereira2019fast}
Talmo~D Pereira, Diego~E Aldarondo, Lindsay Willmore, Mikhail Kislin, Samuel S-H Wang, Mala Murthy, and Joshua~W Shaevitz.
\newblock Fast animal pose estimation using deep neural networks.
\newblock \emph{Nature methods}, 16\penalty0 (1):\penalty0 117--125, 2019.

\bibitem[Pereira et~al.(2022)Pereira, Tabris, Matsliah, Turner, Li, Ravindranath, Papadoyannis, Normand, Deutsch, Wang, et~al.]{pereira2022sleap}
Talmo~D Pereira, Nathaniel Tabris, Arie Matsliah, David~M Turner, Junyu Li, Shruthi Ravindranath, Eleni~S Papadoyannis, Edna Normand, David~S Deutsch, Z~Yan Wang, et~al.
\newblock Sleap: A deep learning system for multi-animal pose tracking.
\newblock 19\penalty0 (4):\penalty0 486--495, 2022.

\bibitem[Radford et~al.(2021)Radford, Kim, Hallacy, Ramesh, Goh, Agarwal, Sastry, Askell, Mishkin, Clark, et~al.]{radford2021clip}
Alec Radford, Jong~Wook Kim, Chris Hallacy, Aditya Ramesh, Gabriel Goh, Sandhini Agarwal, Girish Sastry, Amanda Askell, Pamela Mishkin, Jack Clark, et~al.
\newblock Learning transferable visual models from natural language supervision.
\newblock In \emph{ICML}, pages 8748--8763, 2021.

\bibitem[Rao et~al.(2022{\natexlab{a}})Rao, Xu, Song, Feng, and Wu]{rao2022kitpose}
Jiyong Rao, Tianyang Xu, Xiaoning Song, Zhen-Hua Feng, and Xiao-Jun Wu.
\newblock Kitpose: Keypoint-interactive transformer for animal pose estimation.
\newblock In \emph{PRCV}, pages 660--673, 2022{\natexlab{a}}.

\bibitem[Rao et~al.(2022{\natexlab{b}})Rao, Zhao, Chen, Tang, Zhu, Huang, Zhou, and Lu]{rao2022denseclip}
Yongming Rao, Wenliang Zhao, Guangyi Chen, Yansong Tang, Zheng Zhu, Guan Huang, Jie Zhou, and Jiwen Lu.
\newblock Denseclip: Language-guided dense prediction with context-aware prompting.
\newblock In \emph{CVPR}, pages 18082--18091, 2022{\natexlab{b}}.

\bibitem[Ren et~al.(2024)Ren, Gao, Sun, Qi, Wang, and Liao]{ren2024dynamic}
Pengfei Ren, Yuanyuan Gao, Haifeng Sun, Qi Qi, Jingyu Wang, and Jianxin Liao.
\newblock Dynamic support information mining for category-agnostic pose estimation.
\newblock In \emph{CVPR}, pages 1921--1930, 2024.

\bibitem[Sun et~al.(2019)Sun, Xiao, Liu, and Wang]{sun2019deep}
Ke Sun, Bin Xiao, Dong Liu, and Jingdong Wang.
\newblock Deep high-resolution representation learning for human pose estimation.
\newblock In \emph{CVPR}, pages 5693--5703, 2019.

\bibitem[Wang et~al.(2024)Wang, Zhao, and Chen]{wang2024sql}
Yu Wang, Shengjie Zhao, and Shiwei Chen.
\newblock Sql-net: Semantic query learning for point-supervised temporal action localization.
\newblock \emph{TMM}, 27:\penalty0 84--94, 2024.

\bibitem[Weinreb et~al.(2024)Weinreb, Pearl, Lin, Osman, Zhang, Annapragada, Conlin, Hoffmann, Makowska, Gillis, et~al.]{weinreb2024keypoint}
Caleb Weinreb, Jonah~E Pearl, Sherry Lin, Mohammed Abdal~Monium Osman, Libby Zhang, Sidharth Annapragada, Eli Conlin, Red Hoffmann, Sofia Makowska, Winthrop~F Gillis, et~al.
\newblock Keypoint-moseq: parsing behavior by linking point tracking to pose dynamics.
\newblock \emph{Nature Methods}, 21\penalty0 (7):\penalty0 1329--1339, 2024.

\bibitem[Xiao et~al.(2018)Xiao, Wu, and Wei]{xiao2018simple}
Bin Xiao, Haiping Wu, and Yichen Wei.
\newblock Simple baselines for human pose estimation and tracking.
\newblock In \emph{ECCV}, pages 466--481, 2018.

\bibitem[Xu et~al.(2022{\natexlab{a}})Xu, Jin, Zeng, Liu, Qian, Ouyang, Luo, and Wang]{xu2022pose}
Lumin Xu, Sheng Jin, Wang Zeng, Wentao Liu, Chen Qian, Wanli Ouyang, Ping Luo, and Xiaogang Wang.
\newblock Pose for everything: Towards category-agnostic pose estimation.
\newblock In \emph{ECCV}, pages 398--416, 2022{\natexlab{a}}.

\bibitem[Xu et~al.(2025)Xu, Rao, Song, Feng, and Wu]{xu2025learning}
Tianyang Xu, Jiyong Rao, Xiaoning Song, Zhenhua Feng, and Xiao-Jun Wu.
\newblock Learning structure-supporting dependencies via keypoint interactive transformer for general mammal pose estimation.
\newblock \emph{IJCV}, pages 1--19, 2025.

\bibitem[Xu et~al.(2022{\natexlab{b}})Xu, Zhang, Zhang, and Tao]{xu2022vitpose}
Yufei Xu, Jing Zhang, Qiming Zhang, and Dacheng Tao.
\newblock Vitpose: Simple vision transformer baselines for human pose estimation.
\newblock In \emph{NeurIPS}, pages 38571--38584, 2022{\natexlab{b}}.

\bibitem[Xu et~al.(2024)Xu, Zhang, Zhang, and Tao]{xu2023vitpose++}
Yufei Xu, Jing Zhang, Qiming Zhang, and Dacheng Tao.
\newblock Vitpose++: Vision transformer for generic body pose estimation.
\newblock \emph{IEEE TPAMI}, 46\penalty0 (2):\penalty0 1212--1230, 2024.

\bibitem[Yang et~al.(2024)Yang, Zeng, Zhang, and Zhang]{yang2025xpose}
Jie Yang, Ailing Zeng, Ruimao Zhang, and Lei Zhang.
\newblock X-pose: Detecting any keypoints.
\newblock In \emph{ECCV}, pages 249--268, 2024.

\bibitem[Yang et~al.(2021)Yang, Quan, Nie, and Yang]{yang2021transpose}
Sen Yang, Zhibin Quan, Mu Nie, and Wankou Yang.
\newblock Transpose: Keypoint localization via transformer.
\newblock In \emph{ICCV}, pages 11802--11812, 2021.

\bibitem[Yang et~al.(2023)Yang, Zeng, Yuan, and Li]{yang2023effective}
Zhendong Yang, Ailing Zeng, Chun Yuan, and Yu Li.
\newblock Effective whole-body pose estimation with two-stages distillation.
\newblock In \emph{ICCV}, pages 4210--4220, 2023.

\bibitem[Ye et~al.(2024)Ye, Filippova, Lauer, Schneider, Vidal, Qiu, Mathis, and Mathis]{ye2024superanimal}
Shaokai Ye, Anastasiia Filippova, Jessy Lauer, Steffen Schneider, Maxime Vidal, Tian Qiu, Alexander Mathis, and Mackenzie~Weygandt Mathis.
\newblock Superanimal pretrained pose estimation models for behavioral analysis.
\newblock 15\penalty0 (1):\penalty0 5165, 2024.

\bibitem[Yu et~al.(2021)Yu, Xu, Zhang, Zhao, Guan, and Tao]{yu2021apk}
Hang Yu, Yufei Xu, Jing Zhang, Wei Zhao, Ziyu Guan, and Dacheng Tao.
\newblock {AP}-10k: A benchmark for animal pose estimation in the wild.
\newblock In \emph{NeurIPS}, 2021.

\bibitem[Yuan et~al.(2021)Yuan, Fu, Huang, Lin, Zhang, Chen, and Wang]{yuan2021hrformer}
Yuhui Yuan, Rao Fu, Lang Huang, Weihong Lin, Chao Zhang, Xilin Chen, and Jingdong Wang.
\newblock Hrformer: High-resolution vision transformer for dense predict.
\newblock In \emph{NeurIPS}, pages 7281--7293, 2021.

\bibitem[Zeng et~al.(2024)Zeng, Zhu, Li, Zhao, Shen, and Tang]{zeng2024towards}
Dan Zeng, Yu Zhu, Shuiwang Li, Qijun Zhao, Qiaomu Shen, and Bo Tang.
\newblock Towards labeling-free fine-grained animal pose estimation.
\newblock In \emph{ACMMM}, pages 2545--2553, 2024.

\bibitem[Zhang et~al.(2020)Zhang, Zhu, Dai, Ye, and Zhu]{zhang2020distribution}
Feng Zhang, Xiatian Zhu, Hanbin Dai, Mao Ye, and Ce Zhu.
\newblock Distribution-aware coordinate representation for human pose estimation.
\newblock In \emph{CVPR}, pages 7093--7102, 2020.

\bibitem[Zhang et~al.(2023)Zhang, Wang, Chen, Xu, Zhang, and Tao]{zhang2023clamp}
Xu Zhang, Wen Wang, Zhe Chen, Yufei Xu, Jing Zhang, and Dacheng Tao.
\newblock Clamp: Prompt-based contrastive learning for connecting language and animal pose.
\newblock In \emph{CVPR}, pages 23272--23281, 2023.

\bibitem[Zhou et~al.(2022{\natexlab{a}})Zhou, Yang, Loy, and Liu]{zhou2022conditional}
Kaiyang Zhou, Jingkang Yang, Chen~Change Loy, and Ziwei Liu.
\newblock Conditional prompt learning for vision-language models.
\newblock In \emph{CVPR}, pages 16816--16825, 2022{\natexlab{a}}.

\bibitem[Zhou et~al.(2022{\natexlab{b}})Zhou, Yang, Loy, and Liu]{zhou2022learning}
Kaiyang Zhou, Jingkang Yang, Chen~Change Loy, and Ziwei Liu.
\newblock Learning to prompt for vision-language models.
\newblock \emph{IJCV}, 130\penalty0 (9):\penalty0 2337--2348, 2022{\natexlab{b}}.

\end{thebibliography}
}


\end{document}